# Variable Augmentation Network for Invertible MR Coil Compression

Xianghao Liao, Shanshan Wang, *Senior Member, IEEE*, Lanlan Tu, Yuhao Wang, *Senior Member, IEEE*, Dong Liang, *Senior Member, IEEE*, Qiegen Liu, *Senior Member, IEEE*

*Abstract*—A large number of coils are able to provide enhanced signal-to-noise ratio and improve imaging performance in parallel imaging. Nevertheless, the increasing growth of coil number simultaneously aggravates the drawbacks of data storage and reconstruction speed, especially in some iterative reconstructions. Coil compression addresses these issues by generating fewer virtual coils. In this work, a novel variable augmentation network for invertible coil compression termed VAN-ICC is presented. It utilizes inherent reversibility of normalizing flow-based models for high-precision compression and invertible recovery. By employing the variable augmentation technology to image/k-space variables from multi-coils, VAN-ICC trains invertible networks by finding an invertible and bijective function, which can map the original data to the compressed counterpart and vice versa. Experiments conducted on both fully-sampled and under-sampled data verified the effectiveness and flexibility of VAN-ICC. Quantitative and qualitative comparisons with traditional non-deep learning-based approaches demonstrated that VAN-ICC can carry much higher compression effects. Additionally, its performance is not susceptible to different number of virtual coils.

## I. INTRODUCTION

Parallel imaging (PI) has been proposed to accelerate data acquisitions in magnetic resonance imaging (MRI) during the past two decades [1, 2]. It employs receiver arrays with multiple coils to simultaneously acquire data [1-4]. Different coil sensitivities of the receiver coils are applied to partially replace traditional k-space encoding and reduce scan time. Powered by the rapid development of PI, MRI data acquisitions have been significantly accelerated. This has caused many lengthy MR exams clinically feasible. In addition to PI, there is a growing trend towards utilizing constrained reconstructions in MRI, such as compressed sensing (CS) [5]. However, these reconstructions are frequently iterative and result in computationally intensive. To tackle this deficiency, higher accelerations in data acquisition can be achieved by combining CS with PI [6-8]. Coil arrays are essential for PI methods and the quality of the PI results lies on the number and arrangement of the independent coil elements [3]. Redundancy coil arrays are able to provide high signal-to-noise ratio (SNR) and enhance PI performance. With the development of hardware technology, many researches apply large coil arrays with up to even 128 independent receiver channels [4-7]. As the number of receiver channels increases, nevertheless, the drawbacks of MRI reconstruction speed and data storage in these large datasets are becoming more prominent.

In order to reduce the computational cost for large coil arrays and improve data processing efficiency in the large array systems, the method of coil compression has been proposed [8-11]. The essence of coil channel compression technology is to linearly combine raw data from multiple coil channels into specific virtual coil channel data or to select the most considerable set of data from all virtual coil channel data to reduce the number of coil channel data received independently before the image reconstruction is totally completed [8]. On the one hand, it is capable of significantly reducing the reconstruction time, because the computation amount for constrained and PI reconstruction is determined by the number of coil channels. On the other hand, it can reduce the number of virtual coil channels while retaining as many available signals as possible.

In general, coil compression methods can be classified as the kinds of hardware and software compression. Coil channel compression was initially applied to hardware. Concretely, using the theoretical knowledge of the noise covariance of the receiver array, the original multiple original coils are linearly combined into fewer eigen coils. King *et al.* [11, 12] implemented an efficient coil compression method by combining coil array elements (in hardware) into an alternative basis set with zero noise correlation between array elements. This SNR-preserving data compression method advantageously allows users of MRI systems with fewer receiver channels to achieve the SNR of higher-channel MRI systems. Although the hardware implementation has SNR benefit, it is not always optimal due to lack of consideration for the spatial coil sensitivity variation or the received data, hence the hardware compression causes certain losses and we can't obtain the best image signal.

Compared with hardware compression, software compression is more flexible and results in fewer losses. Buehrer *et al.* [9] proposed a means that achieving coil compression by optimizing SNR of the region of interest in the reconstruction image. It not only depends on the sensitivity of the coil and the accuracy of the sensitivity map, but also requires undersampling to produce a simple point spread function for practical points. Therefore, it is merely suitable for sensitivity-based PI reconstructions such as SENSE [1]. Huang *et al.* [8] developed a different approach to reducing the size of PI data by employing principal component analysis (PCA) in the k-space domain, avoiding the need for coil sensitivities and noise covariance [13-15]. Feng *et al.* [16] presented a method to reduce the computation cost in the k-space domain PI, by utilizing the fact that in large arrays the channel sensitivity is localized to calculate the cross-channel correlation for

This work was supported in part by the National Natural Science Foundation of China under 61871206, 62122033.

X. Liao, L. Tu, Y. Wang and Q. Liu are with the Department of Electronic Information Engineering, Nanchang University, Nanchang 330031, China. ({liaoxianghao, lanlantu}@email.ncu.edu.cn, {wangyuhao, liuqiegen}@ncu.edu.cn)

S. Wang and D. Liang are with Paul C. Lauterbur Research Center for Biomedical Imaging, SIAT, Chinese Academy of Sciences, Shenzhen 518055, China. (sophiasswang@hotmail.com, dong.liang@siat.ac.cn)

channel selection. Zhang *et al.* [3] performed separately for each location along the fully sampled dimensions, which was based on a singular value decomposition (SVD), to compress data in a hybrid image-k-domain. Beatty *et al.* [17] were devoted to a new approach that combined the k-space reconstruction kernel with a coil compression kernel, which was similar to generalized auto calibrating partially parallel acquisitions-based simultaneous-multi slice-acquired coil compression (GRABSMACC), because its unaliasing process was also responsible for coil compression. Doneva *et al.* [18] raised an innovation of correlation with coil sensitivity information, in which the optimal subset of receiving coil channels is determined by the contribution of each coil channel to the final reconstruction. Liu *et al.* [19, 20] took sparse magnetic resonance into consideration, combining the SENSE algorithm in the meanwhile.

The single coil channel compression algorithm (SCC) and the geometric decomposition coil channel compression algorithm (GCC) are both traditional software compression algorithms, which are of high reference value with a long history [3, 8]. Both SCC and GCC require two main steps to achieve coil compression. The first step is to find a suitable coil compression matrix, and the second step is to apply the coil compression matrix to the original coil and construct a virtual coil. Besides, they are able to obtain image quality comparable to that are acquired from all original coil channel data reconstruction. In the same situation, all indexes of GCC are better than those in SCC. The compression loss of GCC is smaller than that of SCC. To achieve similar errors as in GCC, more virtual coils are often demanded in SCC.

Following the success of deep learning in a wide range of applications, neural network-based machine learning techniques have received interest as a means of accelerating MRI. Inspired by the prior arts, for the first time, we employ a deep learning algorithm for MR coil compression. In this work, a variable augmentation network for invertible coil compression termed VAN-ICC is presented for high-precision compression and invertible recovery, which utilizes inherent reversibility of normalizing flow-based models. By applying the variable augmentation technology to image/k-space variables from multi-coils, VAN-ICC trains the invertible network by finding an invertible and bijective function, which can map the original data to the compressed counterpart and vice versa. The whole VAN-ICC can be employed in both image domain and k-space domain, formed VAN-ICC-I and VAN-ICC-K, respectively. More specifically, VAN-ICC-I is a preliminary method to perform coil compression in the image domain. The VAN-ICC-K method is a qualitative leap forward from the VAN-ICC-I method in that it shifts the compression scenario from the image domain to k-space measure, which is more realistic and feasible for hospitals when collecting data, increasing the possibility of applying the method to real-life situations.

The major contributions of this study are as follows:
- This is the first work that presents the idea of invertible coil compression in MR imaging community. Invertible networks are developed for high-precision coil compression and invertible recovery.
- Empowered by the variable augmentation technology and different objective functions, VAN-ICC can be conducted in both image domain and k-space domain, which maps the original data to the compressed counterpart and vice versa.
- The proposed VAN-ICC exhibits great potential application flexibility, which can be conducted on both fully-sampled and under-sampled data. Therefore, it is suitable for the scenarios of effective transport for saving data storage and partially parallel imaging for accelerating reconstruction speed.

The rest of this paper is presented as follows. In Section II, we briefly describe some relevant works on coil compression and invertible model. The proposed VAN-ICC and its main components are introduced in Section III. Section IV plays a critical role in conducting the major experiments. Section V concludes with topics and gives the future work. The work opens a brand new opportunity for invertible MR coil compression without overheads seen in clinical practice.

## II. RELATED WORK

### A. Overview of Coil Compression

Before reconstructing the image, $N < M$ with a series of linear combinations are performed in the time domain to reduce the amount of independent data flowing from the $M$ coil input channels to the $N$ coil output channels, where coil compression comes in. The following equation accurately represents the coil compression problem:

$$x'(k) = Ax(k) \qquad (1)$$

where $x(k)$ denotes a vector that encompasses the data from all $N$ coils at k-space location $k$. $A$ is described as a matrix with dimension $M \times N$ and it is able to be solved by PCA in data-based coil compression [8]. $x'(k)$ demonstrates the data vector from virtual coils that the size reduces from $M$ to $N$.

An approximate representation of the original coil can be achieved by a proper linear combination of virtual coils:

$$x(k) = A^H x'(k) + \varepsilon \qquad (2)$$

where $A^H$ is conjugate transpose matrix of $A$, and $\varepsilon$ stands for the approximate error.

How to minimize the coil compression error with a fixed number of virtual coils is the essence of the coil compression task, i.e.,

$$\min_A \sum_k \left\| (A^H A - I)x(k) \right\|^2 \qquad (3)$$
$$s.t. \quad AA^H = I$$

The solution for Eq. (3) can also be obtained by PCA or SVD. Although many approaches have been devoted to handling Eq. (3), they are lossy compression and the process is not invertible.

### B. Invertible Neural Networks

Varying from the coil compression method in the previous section, the neural network usually maps the input nonlinearly. Nowadays, one of the popular choices in solving image generation tasks is the flow-based invertible network, whose network structure is reversible in principle. Because its inverse process is computationally expensive, it cannot really be used in our daily lives [21-24]. Precisely, NICE and RealNVP present that using coupling layers can make fully invertible networks a natural choice [21, 23]. NICE is the first learning-based normalizing flow framework with the proposed additive coupling layers [23]. RealNVP combines additive and multiplicative coupling layers together to be a general "affine coupling layer", and successfully introduces a convolutional layer in the coupling model, which makes it possible to better deal with image data [21]. The Glow algorithm proposed by Kingma and Dinh [22] incorporates invertible 1×1 convolutions into the affine coupling layer and achieves linearly combine dimensions of the data. Flow++ uses the logistics mixture CDF (cumulative distribution function) coupling flows instead of affine coupling layer and employed purely convolutional conditioning networks in coupling layers [24]. A visualization example of the invertible network is detailed in Fig. 1.

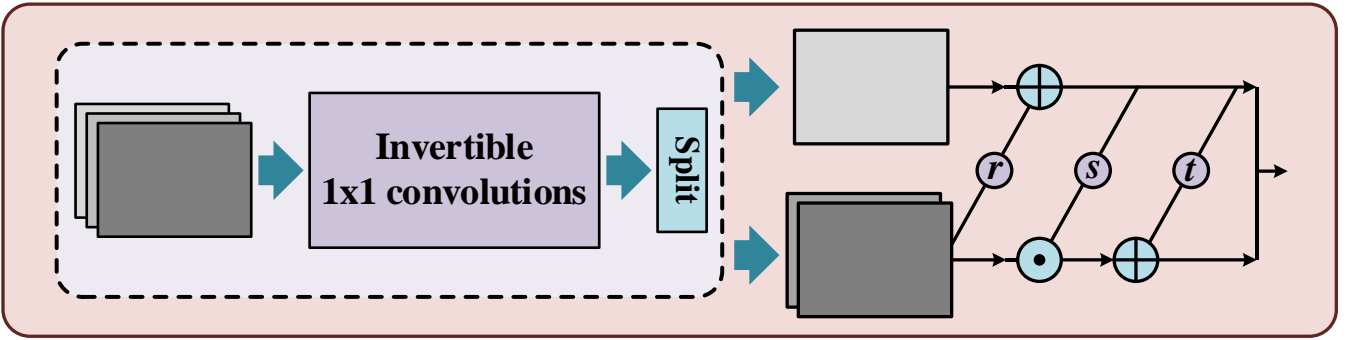

**Fig. 1.** The structure of the invertible block is composed of invertible 1×1 convolutions and an affine coupling layer. $r$, $s$ and $t$ are the transformations defined in the bijective functions $\{f_i\}_{i=0}^{k}$.

*1) Affine Coupling Layer:* Invertible neural networks are typically constructed using coupling layers [21-23]. NICE and RealNVP present coupling layers with a normalizing flow by stacking a sequence of invertible bijective transformation functions [21, 23]. The bijective function is designed as an affine coupling layer, which is a tractable form of Jacobian determinant, and both forward and inverse functions are efficiently computable. In our case, we use an affine coupling layer, which mainly includes three steps [21].

Firstly, it will be more beneficial for us to train deeper networks in the Glow architecture, since each bionic coupling layer performs a specific identity function in its initial state [22]. Secondly, the input involves all the channels and performs a concatenation and segmentation operation. It is worth noting that the segmentation function along the channel dimension first splits the input tensor into two and then concatenates them both into one tensor using latent variables of different scales. In this step, we choose to split the input tensor into three parts of exactly equal size along the channel dimension. Finally, before each step, we perform a reversible convolution, which is done to improve the influence of each dimension on the other dimensions after enough flow steps.

*2) Invertible 1×1 Convolution:* As stated in Glow [22], an invertible 1×1 convolution is added before each affine coupling layer in order to better fuse the information carried by each part after segmentation and to prevent the problem that some channels may never be updated [22]. It is worth noting that a single invertible 1×1 convolution operation is performed, that is, channels with the same number of input and output channels are swapped.

The log-determinant of an invertible 1×1 convolution of a $h \times w \times c$ tensor $h$ with $c \times c$ weight matrix $W$ is straightforward to compute:

$$\log \left| \det\left( \frac{dconv2D(h;W)}{dh} \right) \right| = h \cdot w \cdot \log |\det(W)| \quad (4)$$

The cost of computing of differentiating $\det(W)$ is $O(c^3)$, which is often comparable to the cost of computation $dconv2D(h;W)$ which is $O(h \cdot w \cdot c^3)$. We initialize the weights $W$ as a random rotation matrix, having a log-determinant of 0; after one SGD step these values start to diverge from 0.

Thus, to make the Glow algorithm fast, LU decomposition is used to calculate the matrix mentioned above. The cost of computing $\det(W)$ can be reduced from $O(c^3)$ to $O(c)$ by parameterizing $W$ directly in its LU decomposition:

$$W = PL(U + diag(s)) \quad (5)$$

where $P$ is a permutation matrix, $L$ is a lower triangular matrix with ones on the diagonal, $U$ is an upper triangular matrix with zeros on the diagonal, and $s$ is a vector. The log-determinant is then simplified to be:

$$\log|\det(W)| = sum(\log|s|) \quad (6)$$

For $c$, the difference between computational costs will become more significant, although this large difference in wall clock computing time is not reflected in the networks in which we conducted specific experiments.

In this parameterization, we initialize the parameters by first sampling a random rotation matrix $W$, then computing the corresponding value of $P$ (which remains fixed) and the corresponding initial values of $L$, $U$ and $s$ (which are optimized).

### III. PROPOSED MODEL

#### A. Proposed VAN-ICC

The proposed VAN-ICC consists of the forward process and backward process precisely. We define the forward process of VAN-ICC as mapping MR data of multiple coils to be fewer coils. In addition to the forward process, a reverse process is also needed to restore the hidden features from the compressed data via the VAN-ICC method. In general, in classical neural networks, two completely independent networks are needed to perform these two processes, and this sometimes leads to the production of inaccurate biased mappings. Therefore, we take an alternative measure to implement the reversibility in a single network, which not only simplifies the structure of the network, but also makes training much simpler and easier.

We use the high-dimensional tensor $X$ to denote the original data space and $Y$ to represent the space used to compress the data. It is well known that the MR data or k-space data in our experiments is complex-valued. Theoretically, an $n$-channel image can also be transformed into a $2n$-channel image. Thus, in our experiment, the real and imaginary parts of the $n$-channel MR data are inevitably treated as two separate channels, i.e., $X = \{[x_{real1}, x_{imag1}], \cdots, [x_{realn}, x_{imagn}]\}$. The invertible and bijective function $f: X \rightarrow Y$ maps data points from the original data space to the compressed data space. It is constructed from a sequence of invertible and tractability of Jacobian determinant transformations: $f = f_0 \circ f_1 \circ \cdots \circ f_k$ ($k$ is the number of transformations). Such a sequence of reversible transformations is a normalizing flow. For a given input $x$, by means of a reversible transformation, we can obtain the compressed data $y$ and vice versa

$$y = f_0 \circ f_1 \circ f_2 \circ \cdots \circ f_k (x) \quad (7)$$
$$x = f_0^{-1} \circ f_1^{-1} \circ f_2^{-1} \circ \cdots \circ f_k^{-1} (y) \quad (8)$$

As shown in Fig. 1, the invertible block $f_i$ is implemented through affine coupling layers. In each affine coupling layer, the $D$-dimensional input tensor $u$ is split into two unequal parts, $u_{1:d}$ and $u_{d+1:D}$ ($d < D$), along the channel dimension. Also, $d < D$, the output $v$ can be obtained by

$$v_{1:d} = u_{1:d} \quad (9)$$
$$v_{d+1:D} = u_{d+1:D} \bullet \exp(s(u_{1:d})) + t(u_{1:d}) \quad (10)$$

where $s$ and $t$ represent scale and translation functions. $\bullet$ is the element-wise multiplication. Note that the scale and translation functions are not necessarily invertible, and thus they are realized by a succession of several fully connected layers with leaky ReLU activations.

We reinforce the coupling layer, in order to improve the representation studying ability of the architecture, Eq. (9) is rewritten as

$$v_{1:d} = u_{1:d} + r(u_{d+1:D}) \quad (11)$$

where $r$ can be arbitrarily complicated functions of $u_{d+1:D}$ [25].

Accordingly, given the output $v$, the inverse step is easily expressed as

$$u_{d+1:D} = (v_{d+1:D} - t(v_{1:d})) \bullet \exp(-s(v_{1:d})) \quad (12)$$
$$u_{1:d} = v_{1:d} + r(u_{d+1:D}) \quad (13)$$

Then, in order to replace the reverse permutation operation, which not only reverses the ordering of the channels but also simplifies the architecture, an invertible 1×1 convolution is inserted, presented in [22] between affine coupling layers. It is worth noting that although 3-channel input cannot be even split completely, the invertible 1×1 convolution ensures that unchanged components will be updated promptly in the next invertible block. Moreover, we follow the implementation of [26] and disable batch normalization and weight normalization used in [21, 27].

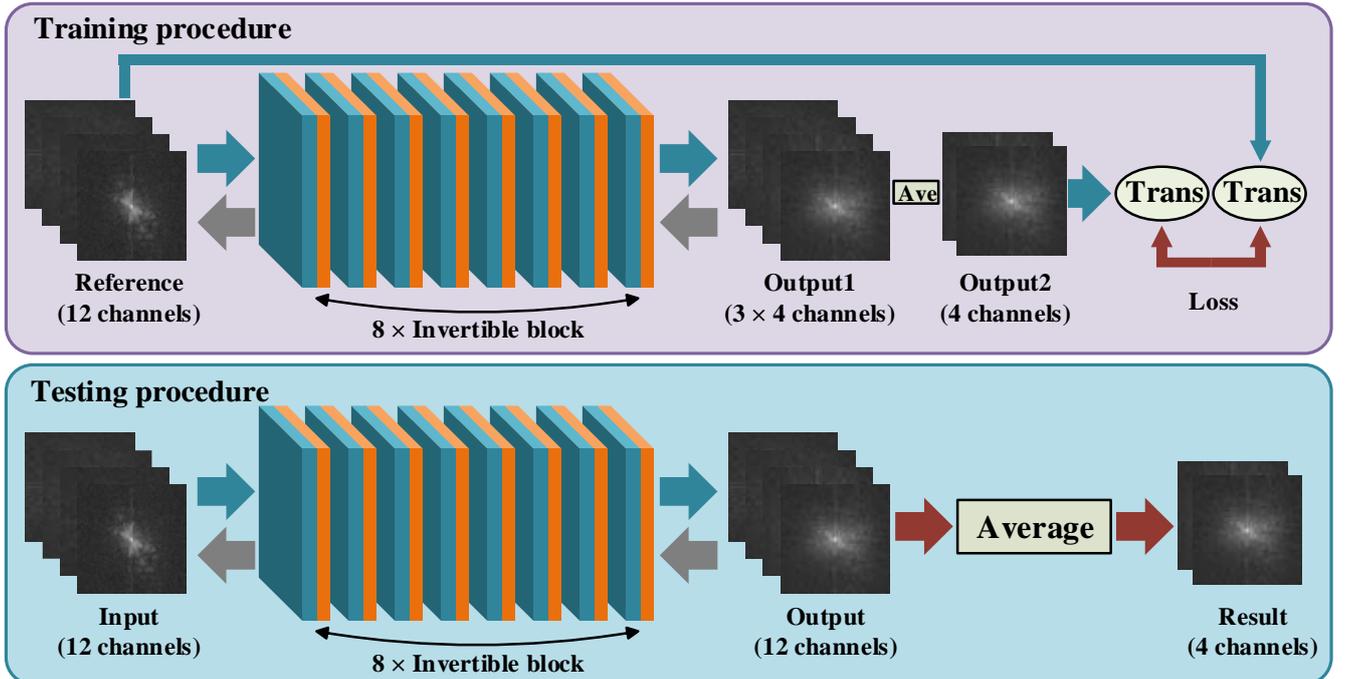

**Fig. 2.** The training and testing procedure of VAN-ICC. Here we take the coil numbers of the input and compressed objects to be 12 and 4 in VAN-ICC-K for example. Average (Ave) denotes the average operator conducted across the channel directions.

The coil compression of VAN-ICC can be conducted in both image domain and k-space domain, forming VAN-ICC-I and VAN-ICC-K, respectively. In Fig. 2, we carry out the 12-channel MR data to be compressed to be a 4-channel object in VAN-ICC-K for example. As can be observed, the input size is equal to the output size in VAN-ICC. Thus, for achieving coil compression, we use an implicated dimension reduction technique by introducing augmented auxiliary variables proposed in [28]. In the training process of this example, a 12-channel output actually consists of three sets of 4-channel data, which can be averaged to obtain a 4-channel object. By means of this variable augmentation, the channel number of the network input and output is same. At the output layer, auxiliary variables aid to provide additional supervision information. After enforcing variable augmentation strategy, the average module enables the demand for dimension reduction to be more accessible. Rather than the VAN-ICC-K that is conducted in k-space domain directly, the implementation of VAN-ICC-I needs a pre-process and a post-process. The compressed MR image can be obtained after sum of squares (SOS) [29-31]. The detailed training process of VAN-ICC-I and VAN-ICC-K is shown in Fig. 3, where the optimization function and network structure will be described in the following subsections.

## B. Optimization Function

Invertible networks can support us to simultaneously minimize the loss in both the network input and output [19], which leads to more effective training results. In the VAN-ICC method, the forward process is utilized to generate 4-channel data and the reverse process is tasked with recovering the original data. It is noteworthy that we employ two completely different loss functions for training forward pass and reverse pass. Moreover, we conduct bi-directional training with smooth $L_1$ loss, which has good overall performance in terms of robustness and stability.

We optimize the VAN-ICC model by minimizing the total loss function

$$L = \lambda L_f + L_r = \lambda \|f(x) - y\|_1^{smooth} + \|f^{-1}(y) - x\|_1^{smooth} \quad (14)$$

where $\lambda$ is the hyper-parameter which is used to balance the accuracy between the compression error and the recovery error. We set $\lambda$ to be 1 in our main experiments. Assume that $d$ represents the error between predicted value and ground truth, the smooth $L_1$ is defined as

$$smooth_{L_1}(d) = \begin{cases} 0.5d^2 & if \ |d| < 1 \\ |d| - 0.5 & otherwise. \end{cases} \quad (15)$$

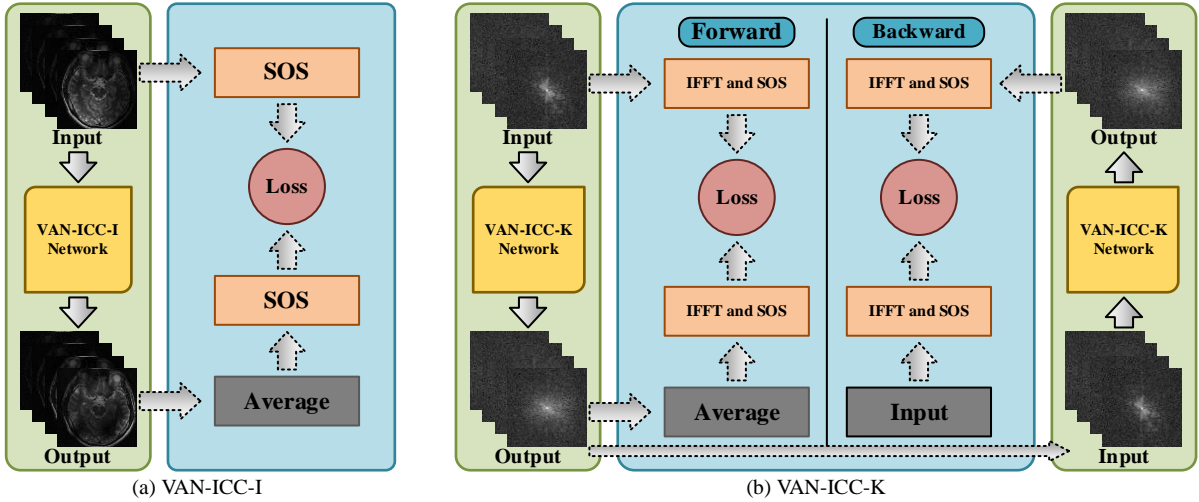

(a) VAN-ICC-I  (b) VAN-ICC-K

**Fig. 3.** The pipeline of the training process of VAN-ICC. Left: The training process of VAN-ICC-I. Right: The training process of VAN-ICC-K. The differences between VAN-ICC-I and VAN-ICC-K are the IFFT and SOS module and that the input and output of VAN-ICC-K are in k-space domain. IFFT stands for the inverse FFT. Average denotes the average operator conducted across the channel directions.

The visualization of the loss functions in VAN-ICC-I and VAN-ICC-K is depicted in Fig. 3(a) and (b), respectively. More precisely, in the image domain of VAN-ICC-I, the loss of forward pass is expressed as followings:

$$L_f = \|sos(x_i) - sos(\hat{x}_i)\|_1^{smooth} \quad (16)$$

Meanwhile, in the k-space domain of VAN-ICC-K, the loss of forward pass is denoted as followings:

$$L_f = \|sos(ifft(x_k)) - sos(ifft(\hat{x}_k))\|_1^{smooth} \quad (17)$$

where the operator $sos$ is defined as $(\sum_{k=1}^{N} |X_k|^2)^{1/2}$, $X_k$ is the complex-valued image from channel $k$, $N$ is the number of channels, $x_i$ and $\hat{x}_i$ denote the ground truth and compressed data in image domain, $x_k$ and $\hat{x}_k$ denote the ground truth and compressed data in k-space domain, $ifft$ is the inverse fast Fourier transform.

The operator $sos$ is applied in image domain. Therefore, we transform the output of VAN-ICC-K to image data. Meanwhile, the input and output of VAN-ICC-K are still in k-space domain. Here, the loss function exhibits two strengths. On one hand, we can get less information loss after compression compared with traditional algorithms. On the other hand, it is unnecessary to collect data with a small number of coils, which reduces the difficulty in data acquisition.

The loss of reverse pass is defined as:

$$L_r = \|x - \hat{x}\|_1^{smooth} \quad (18)$$

where $\hat{x}$ denotes the recovered data.

## C. Network Structure

In this section, the neural network architecture of VAN-ICC will be introduced in detail. The network backbone is called InvBlock, following the Glow which consists of invertible $1 \times 1$ convolution and affine coupling layers [22]. The blocks are depicted pictorially in Fig. 1. In the affine coupling layer, as shown in the right of Fig. 1, the input character is split into two parts along the channel dimension. $r$, $s$, and $t$ are transformations equal to DenseBlock [32], which consists of five 2D convolution layers with filter size $3 \times 3$. Each layer learns a new set of feature maps from the previous layer. The size of the receptive field for the first four convolutional layers is $3 \times 3$, and stride is 2, followed by a Leaky rectified linear unit (LeakyReLU) [33-35]. The last layer is a $3 \times 3$ convolution without LeakyReLU. The purpose of LeakyReLU layers is to refrain from overfitting to the training set and further increase nonlinearity [18, 36, 37, 38].

## IV. EXPERIMENTS

### A. Experiment Setup

*1) Datasets:* In the experiments, we adopt two datasets to evaluate the VAN-ICC-I model and the VAN-ICC-K model. The training datasets are provided by the Shenzhen Institutes of Advanced Technology, the Chinese Academy of Science.

Firstly, the brain data are scanned from a 3T Siemens's MAGNETOM Trio scanner by utilizing the T2-weighted turbo spin-echo sequence. The relevant imaging parameters encompass the size of image acquisition matrix is $256\times256$, echo time (TE) is 149 $ms$, repetition time (TR) is 2500 $ms$, the field of view (FOV) is $220\times220$ $mm^2$ and the slice thickness is 0.86 $mm$. In addition, 480 MR images with 12-channel are employed as training data. In the meanwhile, 20 images are seen as validation data.

Secondly, 101 fully sampled cardiac MR images are acquired with T1-weighted FLASH sequences via taking advantage of 3T MRI scanner (SIEMENS MAGNETOM Trio), whose slice thickness is 6 mm. Typically, the TR/TE is 5/3 $ms$; acquisition matrix size is $192\times192$; FOV is $330\times330$ $mm^2$, and the number of receiver coils is 24. The raw multi-coil data of each frame are combined by an adaptive coil combination method to generate a single-channel complex value image.

*2) Implementation Details:* In this work, we employ MR images as the network input. Notice that the input and output are all complex-valued images with the same size. Each includes real and imaginary components. The smooth $L_1$ loss is adopted for training to compare with the ground truth in the image domain. Adam optimizer is used to train the network. The size of the mini-batch is 1, and the number of epochs in single and multi-coil networks is 1000 and 500, respectively. The initial learning rate is $10^{-5}$, which gradually drops to half until 20 epochs. It is of significant worth to take note of the hyperparameters: $\alpha$ and $\beta$ are the weights associated with different loss terms. The labels for the network are the images generated from direct Fourier inversion from fully sampled k-space data. The sampling in the Fourier domain utilizes the Cartesian sampling strategy, whose accelerated factor is $R=3$.

The proposed method is implemented in python, using Pytorch and Operator Discretization Library (ODL) interface on 2 NVIDIA Titan XP GPUs, 12GB RAM [39]. Training time lasts about 2 days and the source code is available at: *https://github.com/yqx7150/VAN-ICC*.

*3) Evaluation Metrics:* Compression and reconstruction results obtained by VAN-ICC-K are compared relatively to data reconstructed from all 12 coil elements in the original array. To quantitatively evaluate the quality of reconstructed images, we employ three image quality metrics, such as peak signal to noise ratio (PSNR), structural similarity index measure (SSIM) [40-43].

The PSNR is defined by
$$PSNR = 20\log_{10}[Max(x^*)/\|x-x^*\|_2] \quad (19)$$
where $x$ and $x^*$ denote the reconstructed and ground truth images, respectively.

The SSIM index is defined as:
$$SSIM = \frac{(2\mu_x\mu_{x^*}+c_1)(2\sigma_{xx^*}+c_2)}{(\mu_x^2+\mu_{x^*}^2+c_1)(\sigma_x^2+\sigma_{x^*}^2+c_2)} \quad (20)$$
where $\mu_x$ is an average of $x$, $\sigma_x^2$ is a variance of $x$ and $\sigma_{xx^*}$ is a covariance of $x$ and $x^*$. There are two variables to stabilize the division such as $c_1=(k_1L)^2$ and $c_2=(k_2L)^2$. L represents a dynamic range of the pixel intensities. $k_1$ and $k_2$ are constants by default $k_1=0.01$ and $k_2=0.03$.

### B. Experiments on Image Domain

*1) Coil Compression of Fully Sampled Data:* In this experiment, the number of the virtual coils are set to be 4 and 10 for original 12-channel brain images and 20-channel cardiac images, respectively. Quantitative evaluation is placed in Table 1. A comparison of image quality metrics for all compression methods demonstrates the superior performance of VAN-ICC-I. In the brain dataset, the average PSNR value of VAN-ICC-I is higher than SCC and GCC by more than 10.51 dB and 8.68 dB, respectively. Moreover, the SSIM value obtained by VAN-ICC-I is also the highest among the three methods. For the Cardiac dataset, the PSNR value of the introduced method reaches 55.33 dB. However, the SCC and GCC are only 31.15 dB and 36.64 dB. Besides, the SSIM value of VAN-ICC-I is also the highest.

Fig. 4 demonstrates the compression results of SCC, GCC, and VAN-ICC-I in the brain and cardiac datasets. Besides, the SOS images from 12 and 20 original coils are employed as references, respectively. It is obvious that the method of VAN-ICC-I is able to generate realistic compressed images that closely resemble the original ground-truth images. Compared with SCC and GCC, VAN-ICC-I maintains fine features in the edge details. Apart from that, it is most comparable in terms of qualitative visual quality. The outstanding performance can be further indicated through the absolute difference map, unlike the VAN-ICC-I method, both SCC and GCC have lower similarity around the edges or in the background.

TABLE I
QUANTITATIVE RESULTS BY
DIFFERENT METHODS IN THE TEST DATASETS.

| Algorithm | SCC | GCC | VAN-ICC-I |
|---|---|---|---|
| Brain | 41.53/ 0.9808 | 43.36/ 0.9820 | **52.04/ 0.9993** |
| Cardiac | 31.15/ 0.9760 | 36.64/ 0.9917 | **55.33/ 0.9995** |

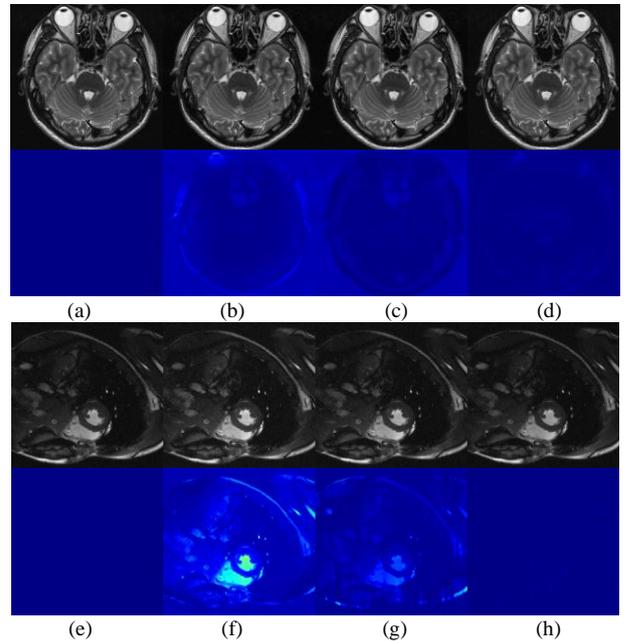

Fig. 4. Comparison of compression results on brain dataset and cardiac dataset for different methods. (Top 2 rows: Brain images, and bottom 2 rows: Cardiac images) Top: (a) Reference image of brain dataset; (b) SCC-4ch; (c) GCC-4ch; (d) VAN-ICC-I-4ch; (e) Reference image of cardiac dataset; (f) SCC-10ch; (g) GCC-10ch; (h) VAN-ICC-I-10ch. Bottom: The $3\times$ absolute difference images between the reference images and compressed images of the corresponding position.

*2) Compression for Different Virtual Coils:* To verify the robustness of the raised VAN-ICC-I, 12-channels brain images are chosen to be compressed into 6 and 4 virtual coils, respectively. Generally speaking, the smaller the number of virtual coils is, the worse the compression performance will be. Fortunately, as exhibited in Fig. 5, VAN-ICC-I is able to still achieve superior results in the circumstance of compressing into fewer virtual coils. As illustrated in Table 2, the average PSNR value of the compressed images in GCC with 4 virtual coils decreases by 6 dB in comparison with the 6 virtual coils result. SCC also reduces 3 dB in the same case. By contrast, the PSNR value accomplished by VAN-ICC-I only descends by 0.24 dB, which is still effective frankly.

*3) Reconstruction of Undersampled Data:* Since our model achieves the excellent performance of coil compression on fully-sampled data, in this subsection we apply VAN-ICC-I to compress under-sampled images for accelerating imaging. More precisely, the under-sampled 20-coil cardiac dataset with Cartesian sampling and acceleration factor is utilized as the network input to be compressed into 10-coil under-sampled images. Then the compressed images are transformed into k-space through Fourier transform. In addition, the compressed k-space data obtained by VAN-ICC-I, SCC, GCC, and the original 20-coil under-sampled k-space data are all reconstructed by $L_1$-SPIRiT [46, 47] algorithm.

Quantitative and qualitative evaluations are listed in Table 4 and Fig. 6, respectively. Fig. 6(c) implies that the reconstruction results by $L_1$-SPIRiT with compressed images through SCC have a large loss of details. Although GCC in Fig. 6(d) has excellent reconstruction quality, its performance is still worse than VAN-ICC-I in Fig. 6(e). Table 3 provides the assessment values corresponding to different methods. The PSNR value of VAN-ICC-I is 32.32 dB and the SSIM value is 0.8676, which are both approaching the result achieved by $L_1$-SPIRiT on 20-coils data. In brief, the proposed VAN-ICC-I is capable of fruitfully accelerating reconstruction, which has fewer errors in comparison with other compression methods.

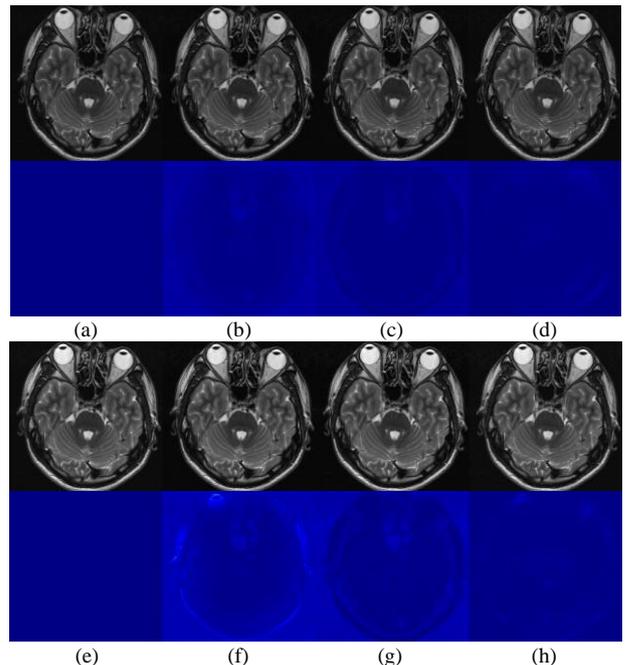

TABLE II
QUANTITATIVE RECONSTRUCTION RESULTS FOR VAN-ICC-I, SCC, AND GCC AT DIFFERENT VIRTUAL COILS.

| Algorithm | SCC | GCC | VAN-ICC-I |
|---|---|---|---|
| 12ch → 6ch | 46.27/0.9915 | 49.47/0.996 | **52.28/0.9995** |
| 12ch → 4ch | 43.53/0.9808 | 43.36/0.982 | **52.04/0.9993** |

**Fig. 5.** Compression results of fully-sampled brain dataset at different virtual coils for different methods. Top: (a)(e) Reference image; (b) SCC-6ch; (c) GCC-6ch; (d) VAN-ICC-I-6ch; (f) SCC-4ch; (g) GCC-4ch; (h) VAN-ICC-I-4ch. Bottom: The absolute difference images between the reference image and compression images.

TABLE III
QUANTITATIVE RECONSTRUCTION RESULTS FOR VAN-ICC-I, SCC, AND GCC ON THE UNDER-SAMPLED DATASET.

| Method | 20ch | SCC-10ch | GCC-10ch | VAN-ICC-10ch |
|---|---|---|---|---|
| Cardiac | 32.49/0.8691 | 29.24/0.8652 | 31.43/0.8657 | **32.32/0.8676** |

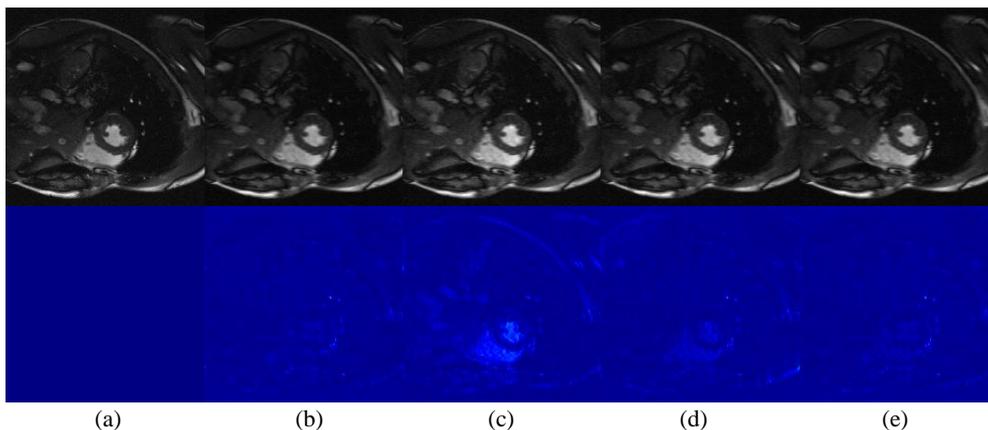

**Fig. 6.** Reconstruction results of the under-sampled cardiac dataset for different methods. Top: (a) Reference image; (b) $L_1$-SPIRiT on 20-coils data; (c) $L_1$-SPIRiT on SCC-10ch data; (d) $L_1$-SPIRiT on GCC-10ch data; (e) $L_1$-SPIRiT on VAN-ICC-I-10ch. Bottom: The absolute difference images between the reference image and reconstruction images.

### C. Experiments on K-space Domain

*1) Coil Compression of Fully Sampled Data:* In contrast to previous experiments with fully sampled data in image domain, in this experiment brain data of 12 channels in k-space domain are compressed into 4 virtual coils and cardiac data of 20 channels in k-space domain are compressed into 10 virtual coils, respectively. The PSNR value and the SSIM value are selected for quantitative assessments of reconstructed image quality in Table 4. For the brain dataset, the PSNR value and the SSIM value of VAN-ICC-K are able to be achieved as excellent indicators. Among the methods, the results compressed by VAN-ICC-K have a good average PSNR value of 47.49 dB which is the highest. VAN-ICC-K also acquires an average SSIM value of 0.9936, which performs well. For the cardiac dataset, the results compressed by VAN-ICC-K reach a mean PSNR value of 40.86 dB, which is superior to SCC and GCC. This method has an average SSIM value of 0.9903, which is competitive compared with other methods.

The compression results are clearly presented in Fig. 7. The reference images are the SOS brain images of 12 original physical coils and the SOS cardiac images of 20 original physical coils. It is striking to observe that the images compressed by VAN-ICC-K have a slight loss and the quality of compression results is almost identical to the original images. In the meantime, the compressed results by VAN-ICC-K maintain features that preserve good imaging details and edge parts, so that the compression images with superb visual effects still retain as much intact and structural information as possible. It is commendably verified by the absolute difference map in Fig. 7.

*2) Compression for Different Virtual Coils:* The following experiment validates the robustness of the previously introduced VAN-ICC-K. For better validation, the experiment chooses to compress the 12 original physical coils brain images into 3, 4, 5, and 6 virtual coils and compress the 20 original physical coils cardiac images into 4, 5, 8, and 10 virtual coils respectively. In general, the compression performance increases with the number of virtual coils in the same case, with a proportional relationship between these two variables. However, it is worth noting that the compression results obtained by the VAN-ICC-K method are not poor, as can be noticed in Fig. 8. The original images are able to retain most of the features incomparably when being compressed into 3 and 4 virtual coils. The quantitative results of this experiment are presented in detail in Table 5. Compared to VAN-ICC-K, the average PSNR values for GCC and SCC are reduced by 8.09 dB and 6.12 dB, respectively when compressing the original image into 3 virtual coils and 4 virtual coils. In the same case, the average PSNR value of VAN-ICC-K increases, making this algorithm undoubtedly optimal among the three in terms of compression performance.

TABLE IV
QUANTITATIVE RESULTS BY DIFFERENT METHODS IN THE TEST DATASETS.

| Algorithm | SCC | GCC | VAN-ICC-K |
|---|---|---|---|
| Brain | 41.37/0.9768 | 43.11/0.9786 | **47.49/0.9936** |
| Cardiac | 31.13/0.9758 | 36.59/**0.9917** | **40.86**/0.9903 |

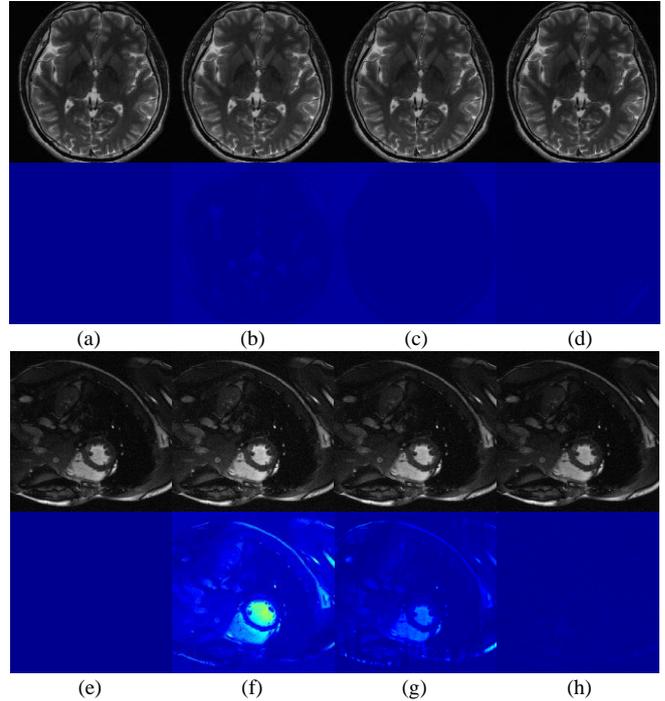

Fig. 7. Comparison of compression results on brain dataset and cardiac dataset for different methods. (Top 2 rows: Brain images, and bottom 2 rows: Cardiac images) Top: (a) Reference image of brain dataset; (b) SCC-4ch; (c) GCC-4ch; (d) VAN-ICC-K-4ch; (e) Reference image of cardiac dataset; (f) SCC-10ch; (g) GCC-10ch; (h) VAN-ICC-K-10ch. Bottom: The 2× absolute difference images between the reference images and compressed images of corresponding position.

TABLE V
QUANTITATIVE RECONSTRUCTION RESULTS FOR VAN-ICC-K, SCC, AND GCC AT DIFFERENT VIRTUAL COILS.

| Algorithm | | SCC | GCC | VAN-ICC-K |
|---|---|---|---|---|
| Brain | 12ch → 3ch | 38.08/0.9763 | 41.36/0.9652 | **46.17/0.9904** |
| | 12ch → 4ch | 41.37/0.9768 | 43.11/0.9786 | **47.49/0.9936** |
| | 12ch → 5ch | 44.07/0.9869 | 45.74/0.9894 | **48.71/0.9952** |
| | 12ch → 6ch | 45.83/0.9897 | 48.64/0.9951 | **48.92/0.9952** |
| Cardiac | 20ch → 4ch | 29.88/0.9207 | 29.77/0.9327 | **34.68/0.9557** |
| | 20ch → 5ch | 30.26/0.9379 | 31.47/0.9516 | **35.13/0.9672** |
| | 20ch → 8ch | 30.87/0.9720 | 34.37/0.9820 | **38.11/0.9822** |
| | 20ch → 10ch | 31.13/0.9758 | 36.59/**0.9917** | **40.86**/0.9903 |

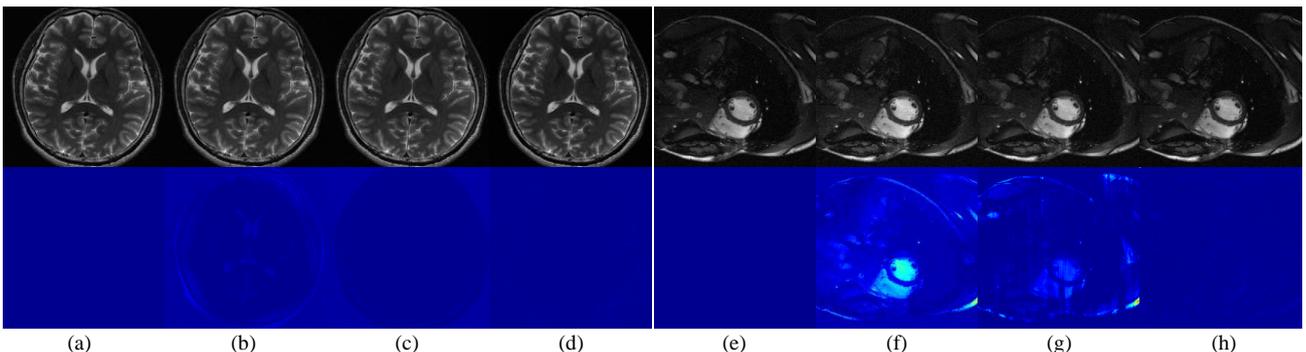

Fig. 8. Compression results of fully-sampled brain dataset at different virtual coils for different methods. (Left 2 rows: Brain images, and right 2 rows: Cardiac images) (a) Reference image of brain dataset; (e) Reference image of cardiac dataset; (b) SCC-3ch; (c) GCC-3ch; (d) VAN-ICC-K-3ch; (f) SCC-8ch; (g) GCC-8ch; (h) VAN-ICC-K-8ch. Bottom: The 2× absolute difference images between the reference images and compression images.

*3) Reconstruction of Under-sampled Data:* In this experiment, we employ VAN-ICC-K to compress under-sampled images to speed up the imaging of the images. To put it in another concrete way, that is the cartesian sampling of the cardiac dataset with an acceleration factor $R=3$. In the first step, the 20-channel under-sampled cardiac MR images in k-space domain are compressed into 10-coil under-sampled images utilizing SCC, GCC, and our proposed VAN-ICC-K methods. In the second and final step, the GRAPPA and $L_1$-SPIRiT [47] algorithms are used to reconstruct the compressed virtual coil images and original physical coil images. The reconstruction results are evaluated quantitatively. Table 6 visualizes the experimental results, the average PSNR values of VAN-ICC-K are 30.30 dB for the GRAPPA algorithm and 30.87 dB for the $L_1$-SPIRiT algorithm, which performs most graceful among the three algorithms. Particularly, the results with virtual coil images compressed by VAN-ICC-K are severely closed to the reconstruction results with original physical coil images. Under the same circumstances, the average SSIM values of VAN-ICC-K are 0.7124 and 0.8104, which are higher than reconstruction results with virtual coil images by SCC and GCC. From Fig. 9, we can accurately find out that the VAN-ICC-K method is implemented to obtain reconstructed images effectively reducing blending artifacts compared with the GCC and SCC methods. In general, VAN-ICC-K has the advantage of minimizing the amount of data storage and computation of reconstruction with slight loss.

TABLE VI
QUANTITATIVE RECONSTRUCTION RESULTS FOR VAN-ICC-K, SCC, AND GCC ON THE UNDER-SAMPLED DATASET.

| Method | 20ch | SCC-10ch | GCC-10ch | VAN-ICC-10ch |
|---|---|---|---|---|
| GRAPPA | 34.13/0.8498 | 29.03/0.7717 | 30.26/0.7950 | **32.04/0.8200** |
| $L_1$-SPIRiT | 31.51/0.8215 | 28.99/0.8074 | 30.11/0.7996 | **30.87/0.8104** |

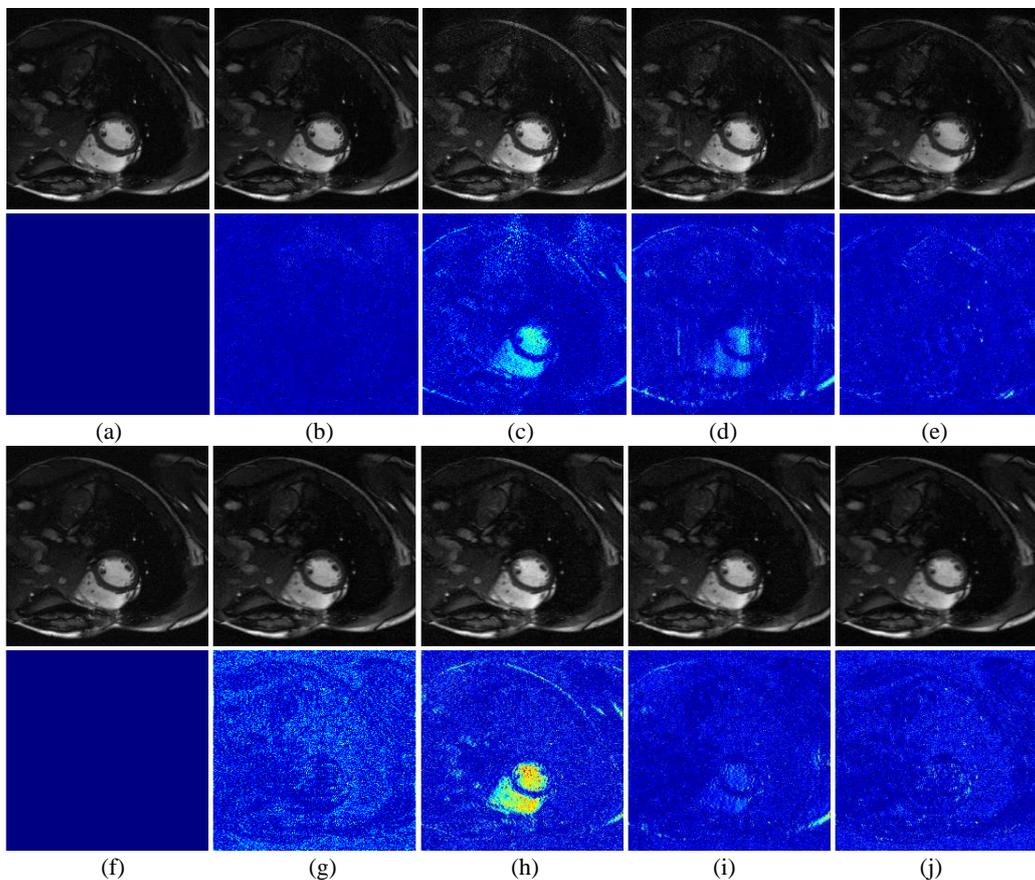

**Fig. 9.** Reconstruction results of under-sampled cardiac dataset for different methods. (Top 2 rows: Reconstruction results for GRAPPA, and bottom 2 rows: Reconstruction results for $L_1$-SPIRiT) Top: (a)(f) Reference image; (b) GRAPPA on original 20-coils data; (c) GRAPPA on SCC-10ch data; (d) GRAPPA on GCC-10ch data; (e) GRAPPA on VAN-ICC-K-10ch; (g) $L_1$-SPIRiT on original 20-coils data; (h) $L_1$-SPIRiT on SCC-10ch data; (i) $L_1$-SPIRiT on GCC-10ch data; (j) $L_1$-SPIRiT on VAN-ICC-K-10ch. Bottom: The 3× absolute difference images between the reference image and reconstruction images.

*4) Time Cost of Compression and Reconstruction:* The greatest advantage of coil channel compression is that it reduces reconstruction time, thus we evaluate the computing efficiency of the coil channel compression technique. The compression time and reconstruction time are the total time taken to process the data. The lower the total time, the more efficient the method. In this experiment, in the first step, we make use of SCC, GCC, and VAN-ICC-K to compress the brain dataset and cardiac dataset into the same virtual coil images. In the second step, the $L_1$-SPIRiT algorithm is utilized to reconstruct the compression results from the first step.

Obviously, during the first step time, our cost is compression time, which is presented in Table 7. In the meanwhile, another reconstruction time is demonstrated in Table 8. From these tables, the compression time and reconstruction time among SCC, GCC, and VAN-ICC-K are barely analogous. Thus, it is not difficult to conclude that VAN-ICC-K efficiently reduces the computational time via compressing the physical coils before reconstructing them, which could improve the computational efficiency and save the image reconstruction time.

*5) Reversible Recovery of Fully Sampled Data:* To verify

the reversibility of the model, the compressed images are taken as the input of the invertible network. By implementing the backward mapping, we can get the restored images with the same number of coils as the original images. The PSNR value and the SSIM value of the forward and backward processes on brain dataset are listed in Table 9. It is observed that the PSNR value of the backward process exceeds 120.68 dB on brain dataset.

Fig. 10 depicts the comparison results of reference images, compressed images and restored images. In comparison, backward process of the proposed VAN-ICC-K generates results that are much closer to the reference images. Overall, by qualitatively and quantitatively comparing the restored results, it can be observed that VAN-ICC-K is able to perform reversible operations on the compressed images, which is not possible for other traditional coil compression methods.

TABLE VII
QUANTITATIVE RESULTS FOR TIME COST OF COMPRESSION.

| Algorithm | SCC | GCC | VAN-ICC-K |
|---|---|---|---|
| Brain | **0.028s** | 0.157s | 0.065s |
| Cardiac | **0.042s** | 0.169s | 0.058s |

TABLE VIII
QUANTITATIVE RESULTS FOR TIME COST OF $L_1$-SPIRIT RECONSTRUCTION.

| Algorithm | Original | SCC | GCC | VAN-ICC-K |
|---|---|---|---|---|
| Brain | 9.9618s | 4.6525s | **4.4172s** | 4.4407s |
| Cardiac | 10.2533s | 5.9394s | 6.1990s | **5.9256s** |

TABLE IX
QUANTITATIVE RESULTS FOR VAN-ICC-K OF FORWARD AND BACKWARD PROCESS.

| | Forward process | Backward process |
|---|---|---|
| Brain | 48.73/0.9945 | 120.68/0.9999 |

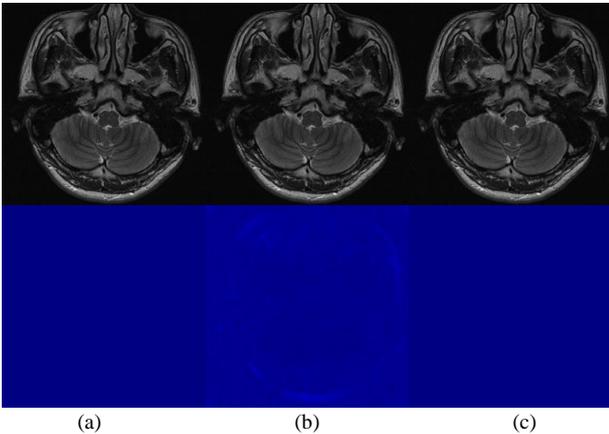

(a)      (b)      (c)

**Fig. 10.** Reversible recovery results of Brain dataset on VAN-ICC-K. Top: (a) Reference image of brain dataset; (b) VAN-ICC-K-6ch; (c) Recovery images by VAN-ICC-K. Bottom: The 5× absolute difference images between the reference image and compression and recovery images.

## V. CONCLUSIONS

In this study, a novel deep learning-based invertible model VAN-ICC was introduced for compressing coils. The present method was built on the invertible neural network as the cornerstone. Auxiliary variables technology was adopted to implicitly reduce the dimension of coils. Therefore, the model was able to not only compress the multi-coils data, but also recovered the compressed image reversibly, which alleviated the burden of data storage and reconstruction time. Experiments indicated that the raised VAN-ICC method outperformed the representative non-deep learning-based approaches, exhibiting great potentials in both research and clinical settings.

REFERENCES

[1] K. P. Pruessman, M. Weiger, M. B. Scheidegger, and P. Boesiger, "SENSE: Sensitivity encoding for fast MRI," *Magn. Reson. Med.*, vol. 42, no. 5, pp. 952-962, 1999.

[2] M. A. Griswold, P. M. Jakob, R. M. Heidemann, M. Nittka, V. Jellus, J. Wang, B. Kiefer, and A. Haase, "Generalized autocalibrating partially parallel acquisitions (GRAPPA)," *Magn. Reson. Med.*, vol. 47, no. 6, pp. 1202-1210, 2002.

[3] T. Zhang, J. M. Pauly, S. S. Vasanawala, and M. Lustig, "Coil compression for accelerated imaging with Cartesian sampling," *Magn. Reson. Med.*, vol. 69, no. 2, pp. 571-582, 2013.

[4] M. A. Ohliger, and D. K. Sodickson, "An introduction to coil array design for parallel MRI," *NMR Biomed.*, vol. 19, no. 3, pp. 300-315, 2006.

[5] K. Nael, M. Fenchel, M. Krishnam, G. Laub, J. P. Finn, and S. G. Ruehm, "High-spatial-resolution whole-body MR angiography with high-acceleration parallel acquisition and 32-channel 3.0-T unit: initial experience," *Radiology*, vol. 242, no. 3, pp. 865–872, 2007.

[6] M. A. Hernandes, R. C. Semelka, J. Elias Júnior, S. Bamrungchart, B. M. Dale, and C. Stallings, "Whole-body MRI: comprehensive evaluation on a 48-channel 3T MRI system in less than 40 minutes. Preliminary results," *Radiologia Brasileira*, vol. 45, no. 6, pp. 319-325, 2012.

[7] C. J. Hardy, R. O. Giaquinto, J. E. Piel, K. W. Rohling, L. Marinelli, D. J. Blezek, E. W. Fiveland, R. D. Darrow, and T. K. F. Foo, "128-channel body MRI with a flexible high-density receiver-coil array," *J. Magn. Reson. Imaging*, vol. 28, no. 5, pp. 1219-1225, 2008.

[8] F. Huang, S. Vijayakumar, Y. Li, S. Hertel, and G. R. Duensing, "A software channel compression technique for faster reconstruction with many channels," *Magn. Reson. Imaging*, vol. 26, no. 1, pp. 133-141, 2008.

[9] M. Buehrer, K. P. Pruessmann, P. Boesiger, and S. Kozerke, "Array compression for MRI with large coil arrays," *Magn. Reson. Med.*, vol. 57, no. 6, pp. 1131-1139, 2007.

[10] M. Doneva, and P. Börnert, "Automatic coil selection for channel reduction in SENSE-based parallel imaging," *Magn. Reson. Mat. Phys. Biol. Med.*, vol. 21, no. 3, pp. 187-196, 2008.

[11] S. B. King, S. M. Varosi, and G. R. Duensing, "Optimum SNR data compression in hardware using an Eigen-coil array," *Magn. Reson. Med.*, vol. 63, no. 5, pp. 1346-1356, 2010.

[12] S. B. King, S. M. Varosi, F. Huang, and G. R. Duensing, "The MRI eigencoil: 2N-channel SNR with N-receivers," *Proc. 11th Annual Meeting of the International Society for Magnetic Resonance in Medicine*, vol. 712, 2003.

[13] M. Ringnér, "What is principal component analysis?," *Nat. Biotechnol.*, vol. 26, no. 3, pp. 303-304, 2008.

[14] G. Destefanis, M. T. Barge, A. Brugiapaglia, and S. Tassone, "The use of principal component analysis (PCA) to characterize beef," *Meat Sci.*, vol. 56, no. 3, pp. 255-259, 2000.

[15] S. Karamizadeh1, S. M. Abdullah, A. A. Manaf, M. Zamani, and A. Hooman, "An overview of principal component analysis," *Journal of Signal and Information Processing*, vol. 4, no. 3B, pp. 173-175, 2013.

[16] S. Feng, Y. Zhu, and J Ji, "Efficient large-array k-domain parallel MRI using channel-by-channel array reduction," *Magn. Reson. Imaging*, vol. 29, no. 2, pp. 209-215, 2011.

[17] P. J. Batty, S. Chang, J. H. Holmes, K. Wang, A. C. S. Brau, S. B. Reeder, and J. H. Brittain, "Design of k-space channel combination kernels and integration with parallel imaging," *Magn. Reson. Med.*, vol. 71, no. 6, pp. 2139-2154, 2014.

[18] A. Radford, L. Metz, and S. Chintala, "Unsupervised representation learning with deep convolutional generative adversarial networks," *arXiv preprint, arXiv:1511.06434*, 2015.

[19] A. Grover, M. Dhar, and S. Ermon, "Flow-gan: Combining maximum likelihood and adversarial learning in generative models," *Thirty-second AAAI conference on artificial intelligence*, 2018.

[20] D. P. Kingma, and J. Ba, "Adam: A method for stochastic optimization," *arXiv preprint, arXiv:1412.6980*, 2014.

[21] L. Dinh, J. Sohl-Dickstein, and S. Bengio, "Density estimation using real nvp," *arXiv preprint, arXiv:1605.08803*, 2016.

[22] D. P. Kingma, and P. Dhariwal, "Glow: generative flow with invertible 1x1 convolutions," *arXiv preprint, arXiv:1807.03039*, 2018.

[23] L. Dinh, D. Krueger, and Y. Bengio, "Nice: Non-linear independent components estimation," *arXiv preprint, arXiv: 1410. 8516*, 2014.

[24] J. Ho, X. Chen, A. Srinivas, Y. Duan, and P. Abbeel, "Flow++: Im-

proving flow-based generative models with variational dequantization and architecture design," *International Conference on Machine Learning. PMLR*, pp. 2722-2730, 2019.

[25] M. Xia, X. Liu, and T Wong, "Invertible grayscale," *ACM Trans. Graph.*, vol. 37, no. 6, pp. 1-10, 2018.

[26] C. Chen, Q. Chen, J. Xu, and V. Koltun, "Learning to see in the dark," *Proceedings of the IEEE Conference on Computer Vision and Pattern Recognition*, pp. 3291-3300, 2018.

[27] S. Ioffe, and C. Szegedy, "Batch normalization: Accelerating deep network training by reducing internal covariate shift," *International conference on machine learning. PMLR*, pp. 448-456, 2015.

[28] Q. Liu, and H. Leung, "Variable augmented neural network for decolorization and multi-exposure fusion," *Inf. Fusion*, vol. 46, pp. 114-127, 2019.

[29] P. Jaini, K. A. Selby, and Y. Yu, "Sum-of-squares polynomial flow," *International Conference on Machine Learning. PMLR*, pp. 3009-3018, 2019.

[30] S. Prajna, A. Papachristodoulou, and P. A. Parrilo, "Introducing SOSTOOLS: A general purpose sum osf squares programming solver," *Proceedings of the 41st IEEE Conference on Decision and Control*, vol. 1, pp. 741-746, 2002.

[31] K. Tanaka, H. Yoshida, H. Ohtake, and H. O. Wang, "A sum-of-squares approach to modeling and control of non-linear dynamical systems with polynomial fuzzy systems," *IEEE Trans. Fuzzy Syst.*, vol. 17, no. 4, pp. 911-922, 2008.

[32] F. Zhang, X. Xu, Z. Xiao, J. Wu, L. Geng, W. Wang, and Y. Liu, "Automated quality classification of colour fundus images based on a modified residual dense block network," *Signal, Image and Video Processing*, vol. 14, no. 1, pp. 215-223, 2020.

[33] A. F. Agarap, "Deep learning using rectified linear units (relu)," *arXiv preprint, arXiv: 1803. 08375*, 2018.

[34] L. Weng, H. Zhang, H. Chen, Z. Song, C. Hsieh, L. Daniel, D. Boning, and I. Dhillon, "Towards fast computation of certified robustness for ReLU networks," *International Conference on Machine Learning. PMLR*, pp. 5276-5285, 2018.

[35] D. Zou, Y. Cao, D. Zhou, and Q. Gu, "Gradient descent optimizes over-parameterized deep ReLU networks," *Mach. Learn.*, vol. 109, no. 3, pp. 467-492, 2020.

[36] B. Xu, N. Wang, T. Chen, and M. Li, "Empirical evaluation of rectified activations in convolutional network," *arXiv preprint, arXiv: 1505. 00853*, 2015.

[37] Y. Liu, X. Wang, L. Wang, and D. Liu, "A modified leaky ReLU scheme (MLRS) for topology optimization with multiple materials," *Appl. Math. Comput.*, vol. 352, pp. 188-204, 2019.

[38] J. Xu, Z. Li, B. Du, M. Zhang, and J. Liu, "Reluplex made more practical: Leaky ReLU," *ISCC*, pp. 1-7, 2020.

[39] G. R. Lee, R. Gommers, F. Waselewski, K. Wohlfahrt, and A. O'Leary, "PyWavelets: A Python package for wavelet analysis," *Journal of Open Source Software*, vol. 4, no. 36, pp. 1237, 2019.

[40] A. Horé, and D. Ziou, "Image quality metrics: PSNR vs. SSIM," *2010 20th international conference on pattern recognition*, pp. 2366-2369, 2010.

[41] Q. Huynh-Thu, and M. Ghanbari, "Scope of validity of PSNR in image/video quality assessment," *Electron. Lett.*, vol. 44, no. 13, pp. 800-801, 2008.

[42] D. Lenyel, D. Kalen, and S. Wolf, "US coast guard high frequency emergency network (HFEN) implementation plan," *NASA STI/Recon Technical Report N*, vol. 85, pp. 29141, 1985.

[43] Z. Chen, H. Lu, S. Tian, J. Qiu, T. Kamiya, S. Serikawa, and L. Xu, "Construction of a hierarchical feature enhancement network and its application in fault recognition," *IEEE Trans. Ind. Inform.*, vol. 17, no. 7, pp. 4827-4836, 2020.

[44] S. Ravishankar, and Y. Bresler, "MR image reconstruction from highly undersampled k-space data by dictionary learning," *IEEE Trans. Ind. Inform.*, vol. 30, no. 5, pp. 1028-1041, 2010.

[45] S. G. Lingala, and M. Jacob, "Blind compressive sensing dynamic MRI," *IEEE Trans. Med. Imaging*, vol. 32, no. 6, pp. 1132-1145, 2013.

[46] M. Lustig, and J. M. Pauly, "SPIRiT: Iterative self-consistent parallel imaging reconstruction from arbitrary k-space," *Magn. Reson. Med.*, vol.64, no. 2, pp. 457-471, 2010.

[47] M. Murphy, K. Keutzer, S. Vasanawala, and M. Lustig, "Clinically feasible reconstruction time for L1-SPIRiT parallel imaging and compressed sensing MRI," *In Proceedings of the ISMRM Scientific Meeting & Exhibition*, vol. 4854, 2010.